\documentclass{bmvc2k}
\pdfoutput=1
\usepackage{multirow}
\usepackage{amsmath}
\usepackage{amssymb}
\usepackage{pifont}
\usepackage{array}
\usepackage{paralist}

\title{Learning Tri-modal Embeddings for Zero-Shot Soundscape Mapping}

\addauthor{Subash Khanal}{k.subash@wustl.edu}{1}
\addauthor{Srikumar Sastry}{s.sastry@wustl.edu}{1}
\addauthor{Aayush Dhakal}{a.dhakal@wustl.edu}{1}
\addauthor{Nathan Jacobs}{jacobsn@wustl.edu}{1}

\addinstitution{
 Computer Science \& Engineering\\
 Washington University in St. Louis \\
 St. Louis, MO, USA
}

\runninghead{Khanal, Sastry, Dhakal, Jacobs}{GeoCLAP}

\def\etal{\emph{et al}\bmvaOneDot}

\usepackage{amsmath}

\begin{document}

\maketitle

\begin{abstract}
We focus on the task of soundscape mapping, which involves predicting the most probable sounds that could be perceived at a particular geographic location. We utilise recent state-of-the-art models to encode geotagged audio, a textual description of the audio, and an overhead image of its capture location using contrastive pre-training. The end result is a shared embedding space for the three modalities, which enables the construction of soundscape maps for any geographic region from textual or audio queries. Using the SoundingEarth dataset, we find that our approach significantly outperforms the existing SOTA, with an improvement of image-to-audio Recall@100 from $0.256$ to $0.450$. Our code is available at \url{https://github.com/mvrl/geoclap}.
\end{abstract}

\section{Introduction}
\label{sec:intro}




Sound is one of the fundamental senses that helps us reason about our environment. There exists an intricate relationship between the visual appearance and sound of a location~\cite{gonzalez2023effects,garzon2023relationships}. Learning about the type of sound at a geographic location allows one to understand many high-level concepts of the area. For example, just by hearing the sound of traffic, we can imagine the location to be an urban setting with a rush of cars and people, whereas the sound of sea waves might elicit the beautiful scenery of a beach. 

There have been several studies conducted on different cities around the world attempting to understand human perception of various types of environmental sound~\cite{aramaki2023image,garzon2023relationships,ooi2023araus,picaut2019open,lionello2020systematic,aiello2016chatty}. Moreover, it has been established that there is a strong correlation between the physiological and psychological health of a person and the environmental sound condition they live in~\cite{lercher2023soundscape,cui2022research,radicchi2021sound}. Therefore, understanding the soundscape for a given geographic area can be of great importance to policymakers focused on urban planning and environmental noise management. Soundscapes also serve value to the general public for whom environmental sound plays a vital role in decisions such as buying a house or setting up a business.

Most of the existing works on creating soundscape focus on crowd-sourcing human perception of sound in their surroundings~\cite{aiello2016chatty,lionello2020systematic,picaut2019open,aramaki2023image,yue2023visualized}. While serving as an important tool for understanding the sound distribution of a region, such approaches have two major limitations. First, the abstraction of sound into a fixed set of indicators and psycho-acoustic descriptors limits our ability to have a complete picture of underlying physical factors associated with sound. Second, such soundscapes are usually created for only highly visited places in the world, creating massive sparsity of soundscapes on a global scale. In order to solve both of these limitations, we propose to leverage the intrinsic relationship between sound and visual cues of the location and learn to directly predict the most probable sound that could be heard at any given location. Specifically, we train a multi-modal deep learning framework that learns a shared embedding space where the sound that is most likely to come from a given location, is pulled closer while pushing other unlikely sounds farther apart. We represent the location (latitude, longitude) by an overhead image of size $H \times W$ centered around it. Once trained, our multi-modal embedding space and free availability of overhead imagery makes it possible for us to create soundscape maps for any area in the world.

One of the successful approaches to learning shared embedding space between different modalities is contrastive learning. In recent years, contrastive learning between image and text~\cite{radford2021learning}; image, text, and audio~\cite{guzhov2022audioclip}; text, audio~\cite{elizalde2023clap,deshmukh2022audio,laionclap2023}; overhead image and audio~\cite{heidler2023self} has been an effective self-supervised training objective to learn a multi-modal embedding space. Such a space has an understanding of the correspondence between the modalities that can be transferred to various downstream tasks, where impressive results have been observed. Motivated by these works, we also adopt contrastive learning as our pre-training strategy to learn a multi-modal embedding space. However, unlike the prior works, we are interested in incorporating geographic knowledge into the embedding space learned by audio-language pre-training. We achieve this by adding an overhead image, capturing the geographic context of a scene, as an additional modality in our contrastive learning framework. With the shared embedding space that has knowledge of correspondence between audio and its corresponding overhead image, we can then formulate the task of soundscape mapping as a cross-modal retrieval problem, where the objective is to predict the most likely sound from a gallery of $N$ sounds given an overhead image.

Our work builds upon a prior work~\cite{heidler2023self} that introduced the {\em SoundingEarth} dataset containing over $50k$ geotagged audios paired with their corresponding overhead image. The objective of work by Heidler \etal~\cite{heidler2023self} was to learn a good audio-visual embedding space useful to be transferred for different downstream tasks in remote sensing. However, in the interest of learning an embedding space to create accurate soundscapes, our work is focused on improving the task of cross-modal retrieval. In this regard, we utilise weights of publicly available modality-specific SOTA models. Moreover, unlike Heidler \etal, who build an embedding space capturing two modalities (overhead-image and audio), we propose to also incorporate textual description of audio into the embedding space. This essentially creates a tri-modal embedding space with richer understanding of three modalities: overhead-image, audio, and text. We call our framework GeoCLAP: Geography-Aware Contrastive Language Audio Pre-training. As demonstrated by our results adding the textual modality improves the representational capability of both overhead-image and audio encoders. Moreover, with an understanding of three modalities, we are now able to create soundscapes either from a textual or audio query for any geographic region.  The main contributions of our work are as follows:
\begin{compactitem}
 \item We significantly improve the prior baseline on the task of cross-modal retrieval of overhead image to sound and vice-versa.  
 \item We build a tri-modal embedding space that has an understanding of overhead image, audio, and textual description of audio at a given location.
 \item We demonstrate a simple and scalable way of creating soundscape for any geographic area using either a textual or audio query. 
\end{compactitem}

\section{Related Work}
\label{sec:litreview}
\subsection{Soundscape Mapping}
The soundscape of a geographic region can be defined as the acoustic environment perceived by individuals within its context~\cite{international2014iso}.~There exists a large body of work focusing on the problem of soundscape mapping~\cite{aiello2016chatty,lionello2020systematic,picaut2019open,margaritis2017soundscape,aramaki2023image,engel2021review,gonzalez2023effects,ooi2023araus,zhao2023sensing,yue2023visualized}. In these works, soundscape mapping is formulated as a framework containing three components: indicators, descriptors, and a predictive model that maps indicators to descriptors. Indicators are psycho-acoustic measures (for example, sound pressure level, loudness, spectral slope, etc.) which determine the perceived value of descriptors (for example, pleasant, unpleasant, eventful, etc.). In this paper, we refer to this line of work as perceptual soundscape mapping.

One of the common findings from the literature of perceptual soundscape mapping is that there exists a strong correlation between the human perception of sound and the environmental variables of the scene such as building, road category, etc.~\cite{garzon2023relationships}. Utilising this correlation between sound and visual cues, there have been a few works that use deep learning to learn a shared embedding space between sound and either ground level image~\cite{owens2016ambient} or overhead image~\cite{hu2020cross} of the scene. This multimodal learning approach leads to improved performance on visual tasks such as aerial scene recognition~\cite{hu2020cross}, image classification~\cite{owens2016ambient}, and object detection~\cite{owens2016ambient}. Closer to our work, a few prior works~\cite{mao2018deep,chen2020deep,ning2021semantics, yang2022multimodal} focus on the task of cross-modal image-to-voice retrieval. Such tasks require a dataset containing overhead imagery paired with spoken audio captions, which is very limited. Moreover, instead of learning from speech, we are interested in learning from free-form audio such as field recordings, natural sounds, etc. which capture diverse concepts of the location. Another closer work by Salem \etal~\cite{salem2018multimodal}, proposed learning a shared embedding space between audio, overhead image, and ground level image, enabling them to predict a distribution over sound clusters from an overhead image. The problem formulation of soundscape mapping in our work is similar to~\cite{salem2018multimodal}. However, the striking difference as well as the strength of our work is that leveraging the power of contrastive language audio pre-training~(CLAP), we are able to create  soundscape conditioned on any textual or audio query. In doing so, we still retain the ability to create soundscape with desired set of sound categories in a zero-shot manner. 

\subsection{Contrastive Learning}
Radford \etal, in their seminal work, CLIP~\cite{radford2021learning}, trained large image-text dataset using contrastive loss and demonstrated it's impressive zero-shot performance on many computer vision tasks. AudioCLIP~\cite{guzhov2022audioclip}, extends CLIP to three modalities: image, text, and audio. Such tri-modal embedding space enables one to perform query between three pairs of modalities. Wav2clip~\cite{wu2022wav2clip}, distilled the knowledge of CLIP embedding space by freezing the image encoder of CLIP and contrastively training an audio encoder to learn a new embedding space shared by audio and a corresponding image. With similar training objective as CLIP, another work CLAP~\cite{elizalde2023clap} performs contrastive learning between audio and natural language. CLAP training has proven to be an effective strategy with impressive audio retrieval performance~\cite{deshmukh2022audio}. Inspired by this, Wu \etal~\cite{laionclap2023} further improved the CLAP’s performance by training on large-scale data with effective audio feature fusion and text augmentation strategies. We refer the work by  Wu \etal~\cite{laionclap2023} as L-CLAP in our paper and use the pre-trained encoders from L-CLAP to embed audio and text for GeoCLAP pre-training.

Our work takes motivation from the proven performance of contrastive learning as an effective pre-training strategy. The focus of our work is soundscape mapping. The embedding space for such tasks should have an understanding of geography of a location where the sound is coming from~\cite{ayush2021geography}. Therefore, we propose to learn an embedding space trained contrastively on three modalities: overhead image, text, and audio.

\subsection{Pretrained Models}
Availability of modality specific pre-trained models trained with various self-supervision objectives have proven to be crucial in bringing performance improvement in various tasks in remote sensing~\cite{wang2022self}. In the recent years, masked auto-encoders (MAE)~\cite{he2022masked} based models trained on satellite imagery have demonstrated to be a good starting checkpoints to be fine-tuned for various downstream tasks~\cite{cong2022satmae,reed2022scale}. In our work, we start with the pre-trained weights of Vision Transformer (ViT)~\cite{dosovitskiy2020vit} encoder of SATMAE~\cite{cong2022satmae} as the overhead-image encoder for GeoCLAP. SATMAE~\cite{cong2022satmae} was pre-trained on large-scale (over 700K) satellite imagery of the world.
To learn representations for audio and text, we use L-CLAP's pre-trained encoders. It uses HTSAT~\cite{chen2022hts} as the audio encoder and RoBERTa~\cite{liu2019roberta} as the text encoder. HTSAT is a swin-transformer~\cite{liu2021swin} based model with SOTA performance on various audio classification tasks. RoBERTa is a powerful transformer-based language model trained with improved design choices than BERT~\cite{devlin2018bert}. L-CLAP~\cite{laionclap2023} was contrastively pre-trained on over 630K audio-text paired dataset.

\section{Approach}
\label{sec:approach}

We present a detailed description of our approach, including the high-level problem formulation, a description of our primary evaluation dataset, and a detailed description of the network architecture and training procedure for our method, GeoCLAP.

\subsection{Problem formulation}

The objective of our work is to learn a shared embedding space that allows us to predict the most probable sounds that can be heard at a given geographic location. This can be represented as $s^* = \max_{s}P(s|l)$ where $P(s|l)$ represents the conditional distribution of sounds for a given location $l$ and $s^*$ is the most likely sound. Unfortunately, direct conditioning on location does not generalize to regions without a large number of training samples, which means truly global mapping wouldn't be possible. On the other hand, overhead imagery has a strong correlation to the type of sound at a given location and is freely available across the globe. Therefore, in our work, we represent the location indirectly, using an overhead image $I(l)$ of the location. We learn a conditional distribution $P(s|I(l))$, which is able to make high-resolution predictions even for regions without training samples.

\subsection{Dataset}
\label{sec:dataset}

We use the {\em SoundingEarth} dataset to train and evaluate our method. The dataset contains more than $50k$ geotagged audio recordings from $136$ countries and overhead image pairs. The overhead images have size of  ${1024} \times {1024}$ collected from {\em Google Earth} with an approximate ground-sample distance (GSD) of $0.2$ meters~(m). Audio data in the dataset was collected from the project {\em Radio Aporee:::Maps}~\cite{aporee}, which hosts an online platform dedicated to creating a global soundmap. It contains diverse audio recordings from urban, rural and natural environments, published under the creative commons license. For our project, we remove the audio files with a sampling frequency less than $16k$. This yields a dataset size of $50\,792$ samples. 

The high-resolution~{\em Google Earth} imagery is not available to be used freely. Therefore, in order to have the ability to globally scale soundscape mapping, we augment the existing~{\em SoundingEarth} dataset by including freely available lower-resolution images. Specifically, we use the RGB bands of the \textit{Sentinel-2 cloudless} imagery with $10m$~$GSD$. For each location, we download a $256 \times 256$ image tile with the coverage radius of $512m$ centered at that location.

\subsection{GeoCLAP}

\begin{figure}[!ht]
    \centering
    \includegraphics[scale=0.55]{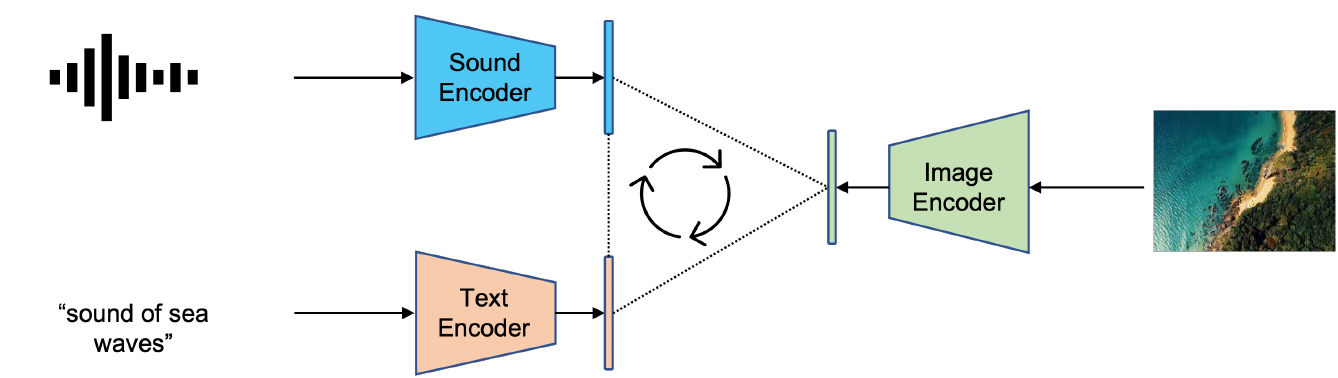}
    \caption{GeoCLAP: A tri-modal contrastive learning framework to learn shared embedding space between overhead image, sound, and textual description of the corresponding sound.}
    \label{fig:geoclap}
\end{figure}

Figure~\ref{fig:geoclap} represents the overall framework of GeoCLAP. Given a geotagged audio~$X_{k}^{a}$, textual description of the audio~$X_{k}^{t}$, and an overhead image at a given location~$X_{k}^{i}$, where ($X_{k}^{a}$,$X_{k}^{t}$,$X_{k}^{i}$) is one audio-text-image triplet. We obtain embeddings for each modalities by passing through modality-specific encoder and linear projection layer, yielding embeddings with the same dimension for audio, text, and overhead image, respectively.

\begin{equation}\label{eq:1}
E_{k}^{a} = g_{audio}(f_{audio}(X_{k}^{a}))
\end{equation}
\begin{equation}\label{eq:2}
E_{k}^{t} = g_{text}(f_{text}(X_{k}^{t}))
\end{equation}
\begin{equation}\label{eq:3}
E_{k}^{i} = g_{image}(f_{image}(X_{k}^{i}))
\end{equation} where $(f_{audio}, g_{audio})$, $(f_{text}, g_{text})$, $(f_{image}, g_{image})$  are (encoder, linear projection layer) pairs producing $l2$-normalized $d$ dimensional embeddings: $E_{k}^{a}$, $E_{k}^{t}$, and $E_{k}^{i}$, for audio, text, and overhead image respectively. 

GeoCLAP is trained on embedding triplets using contrastive learning objective similar to CLIP~\cite{radford2021learning} for all three pairs of embeddings:

\begin{equation}\label{eq:4}
    L_{at} = \frac{1}{2N}\sum_{k=1}^{N}\bigg(log \frac{exp(E_{k}^{a}.E_{k}^{t}/\tau_{at})}{\sum_{j=1}^{N}exp(E_{k}^{a}.E_{j}^{t}/\tau_{at})} + log\frac{exp(E_{k}^{t}.E_{k}^{a}/\tau_{at})}{\sum_{j=1}^{N}exp(E_{k}^{t}.E_{j}^{a}/\tau_{at})}\bigg)
\end{equation}

\begin{equation}\label{eq:5}
    L_{ai} = \frac{1}{2N}\sum_{k=1}^{N}\bigg(log \frac{exp(E_{k}^{a}.E_{k}^{i}/\tau_{ai})}{\sum_{j=1}^{N}exp(E_{k}^{a}.E_{j}^{i}/\tau_{ai})} + log\frac{exp(E_{k}^{i}.E_{k}^{a}/\tau_{ai})}{\sum_{j=1}^{N}exp(E_{k}^{i}.E_{j}^{a}/\tau_{ai})}\bigg)
\end{equation}

\begin{equation}\label{eq:6}
    L_{ti} = \frac{1}{2N}\sum_{k=1}^{N}\bigg(log \frac{exp(E_{k}^{t}.E_{k}^{i}/\tau_{ti})}{\sum_{j=1}^{N}exp(E_{k}^{t}.E_{j}^{i}/\tau_{ti})} + log\frac{exp(E_{k}^{i}.E_{k}^{t}/\tau_{ti})}{\sum_{j=1}^{N}exp(E_{k}^{i}.E_{j}^{t}/\tau_{ti})}\bigg)
\end{equation} where, $N$ is the training batch size and $\tau_{at}$, $\tau_{ai}$, and $\tau_{ti}$ are learnable temperature parameters used to scale logits in loss computation for each pairs of embeddings.

Combining equations \ref{eq:4}, \ref{eq:5}, and \ref{eq:6}, the final loss for which GeoCLAP is trained is as follows:
\begin{equation}\label{eq:7}
    L = L_{at} + L_{ai} + L_{ti}
\end{equation}

\newcommand{\cmark}{\ding{51}}%
\newcommand{\xmark}{\ding{55}}%

\section{Experimental Details}
\subsection{Data Preprocessing}
For audio preprocessing, we convert each audio sample into mel-spectrogram using the default settings:~\texttt{\{feature\_size=64, sampling\_rate=48000, hop\_length=480, max\_length\_s=10, fft\_window\_size=1024\}} provided in the \texttt{HuggingFace}-wrapper:\texttt{ClapProcessor} for the pre-trained L-CLAP model~\texttt{clap-htsat-fused}.

In the {\em SoundingEarth} dataset, most of the audio recordings (except $6333$ samples) are also accompanied by a brief description and a title uploaded by the contributor. In order to have textual description for all audio recordings as well as to further encode geographic information in text, we use a python client,~\texttt{geopy} to obtain the address of the location and append an additional sentence, \emph{``The location of the sound is:\{address\}.''} to the textual description of each sample. For example, for the geolocation $(52.509663, 13.376481)$, the added sentence would be \emph{``The location of the sound is: Potsdamer Platz, Tiergarten, Mitte, Berlin, 10785, Germany''}. Following L-CLAP, we use \texttt{RobertaTokenizer} with the parameter \texttt{max\_length} set to $77$. 

For overhead imagery, we adopt the same data augmentation as SATMAE~\cite{cong2022satmae}. We perform \textit{RandomResizedCrop} with parameters:\texttt{\{input\_size=224, scale=(0.2,1.0)\}}, followed by a \textit{RandomHorizontalFlip}, during training. During inference, we extract a $224\times224$ center crop of the image.

\subsection{Implementation and metrics}
We implement our code in \texttt{Pytorch} and utilise \texttt{HuggingFace} for loading L-CLAP encoders and their respective data pre-processing wrappers.  We split the dataset with ratio $70$:$10$:$20$ yielding a total of $35\,554$, $5\,079$, and $10\,159$ samples into training, validation, and test split, respectively. For experiments regarding the baseline, we ran the publicly available code for~\cite{heidler2023self} using the data splits of our study. We used the experimental setting for their best reported results on cross-modal retrieval task, which is as follows:\texttt{\{batch\_size=256 encoders=ResNet18, latent\_dim=128, loss=SymmetricCL, tau=0.2\}}. The baseline was trained for $300$ \texttt{epochs} with \texttt{Adam} optimizer and learning rate of $1e-3$.

\subsubsection{Encoders}
We use the pre-trained model~\texttt{clap-htsat-fused}~\cite{laionclap2023} to encode audio and text. The audio encoder used in our study, HTSAT, has $4$ swin-transformer blocks with hidden feature dimension of $768$. The text encoder RoBERTa from~\cite{laionclap2023} used in our study, has $12$ transformer blocks with hidden feature dimension of $768$. For both audio and text encoders, we take the output of their respective L-CLAP's projection layer producing $512$-dimensional embeddings.
For encoding overhead image, we use the pre-trained \texttt{vit\_base\_patch16} encoder of SATMAE~\cite{cong2022satmae}. It processes input as a sequence of $16\times16$ image patches passing through $12$ layers of transformer blocks. In order to match dimension with audio and text embeddings, we pass the output from SATMAE encoder to a ReLU activation followed by a $512$-dimension linear layer.
Starting from weights of these pre-trained encoders, we conduct two set of experiments. First, we allow only the overhead-image encoder to train while freezing L-CLAP. Second, we allow fine-tuning of all encoders in our framework. 

\subsubsection{Training}
We train GeoCLAP using the contrastive loss objective presented in Equation~\ref{eq:7}. We initialize all three learnable temperature parameters to $0.07$. We also run experiments with and without using \textit{text} in our framework. While using text, we further experiment the impact of adding an additional sentence describing detailed address of the location to the text. For experiments where we use overhead image and audio only, we train our model with image-audio contrastive loss represented by Equation~\ref{eq:5}. Moreover, for experiments using overhead image, audio, and text, while keeping the L-CLAP encoders frozen, we train with $Loss = L_{ai} + L_{ti}$. We use a training batch size of $256$ for the baseline, and our experiments with frozen L-CLAP, while using batch size of $128$ for experiments allowing fine tuning of L-CLAP. We use the \texttt{Adam} optimizer and set the initial learning rate to $5e-5$. We use \texttt{weight\_decay=0.2} and \texttt{betas=(0.9,0.98)}. We use cosine annealing learning rate scheduler with number of warm up iterations set to $2000$. We set~\texttt{max\_epochs} to $100$ for experiments with frozen L-CLAP and $30$ for experiments allowing fine tuning of L-CLAP. 

\subsubsection{Metrics}
Following Heidler \etal~\cite{heidler2023self}, we use Recall@100 and Median Rank (Median-R) of the ground-truth as the evaluation metrics of our approach. We use the test set containing $10\,159$ samples as the gallery for both image-to-sound and sound-to-image retrieval evaluation.

\section{Evaluation}
\label{sec:evaluation}
\subsection{Experiments with SoundingEarth data}
Table~\ref{table:1} shows the results of our experiments with the {\em SoundingEarth} dataset while using the original overhead imagery of $0.2m$ resolution. One of the interesting results from this table is that by just using frozen pre-trained audio encoder from L-CLAP~\cite{laionclap2023}, while allowing only overhead-image encoder to be fine-tuned, we already get about $10$ points improvement in cross-modal retrieval. This highlights the advantage of leveraging rich representation space of pre-trained models like L-CLAP. However, when we introduce text modality into training, while still keeping both text and audio encoders frozen, the image-to-sound Recall@100 drops to $0.32$. L-CLAP was trained on large corpus of text-audio pairs where textual description of audio have relatively high quality. However, the primary focus of the {\em SoundingEarth} dataset has been to collect geotagged audio from all around the world and associate them with high-resolution overhead imagery. We observed that the textual descriptions of audio in the {\em SoundingEarth} dataset are noisy and do not reflect the type of textual prompts L-CLAP models were trained on. In our experiments, we use three different types of texts: textual description of audio, only address of the audio, and text containing both description and address of the audio. We observed that for any type of text, learning with frozen representation lowers the performance when compared to learning with frozen representation of audio alone. With this observation, we decided to allow fine-tuning of L-CLAP encoders. Accordingly, the performance of our approach noticeably improves to image-to-sound Recall@100 of $0.384$ while learning with overhead image and audio. The performance further improves to Recall@100 of $0.423$ with Median Rank of $172$ when we learn with overhead image, audio, and text. 
This performance is further improved to Recall@100 of $0.434$ with Median Rank of $159$, when we add address of the audio location in the text. This is an absolute improvement of the baseline performance by $0.178$ points in image-to-sound Recall@100 and $655$ in Median Rank.  We see similar trends on sound-to-image retrieval task.
\begin{table}
\scriptsize
 \begin{center}
    \begin{tabular}{ c c c c c|c c|c c  }
 \hline
 \multicolumn{5}{c|}{Method} &
 \multicolumn{2}{c|}{Image2Sound} &
 \multicolumn{2}{c}{Sound2Image} \\
 
 \hline
 Experiment & Image Encoder & Text-Audio Encoder & Text & Address & R@100 & Median-R & R@100 & Median-R\\
 \hline
 Baseline~\cite{heidler2023self} & ResNet18 & ResNet18 & \xmark & \xmark & 0.256 & 814 & 0.250 & 816\\
 \hline
 ours & SATMAE & L-CLAP-frozen& \xmark & \xmark & 0.352 & 360 & 0.348 & 369\\
 ours & SATMAE & L-CLAP-frozen & \cmark & \xmark & 0.328& 428 & 0.325 & 428 \\
 ours & SATMAE & L-CLAP-frozen & \xmark & \cmark & 0.298 & 546 &0.295 & 544 \\
 ours & SATMAE & L-CLAP-frozen & \cmark & \cmark & 0.317 & 439 & 0.311 & 443\\
 \hline
 ours & SATMAE & L-CLAP & \xmark & \xmark & 0.384 & 230 & 0.385 & 237\\
 ours & SATMAE & L-CLAP & \cmark & \xmark &0.423  & 172 & 0.419 & 175 \\
 ours & SATMAE & L-CLAP & \xmark & \cmark & 0.432 & 166 & 0.431 &167\\
 ours & SATMAE & L-CLAP & \cmark & \cmark & \textbf{0.434} & \textbf{159} & \textbf{0.434} & \textbf{167}\\
\hline

\end{tabular}
 \end{center}
 \caption{Cross-modal retrieval performance for models using 0.2m GSD overhead imagery.}
\label{table:1}
\end{table}

\subsection{Experiments with Sentinel data}
Table~\ref{table:2} shows the results of our experiments with \textit{Sentinel-2 cloudless} imagery with $10m$~$GSD$. We found that performance in all of our experiments noticeably improved while using lower-resolution overhead imagery. This choice brought in $12.89\%$ relative improvement in the baseline Recall@100 performance as well. We believe the reason for this improvement is the larger coverage of geographic area in a single overhead image with $10$m~$GSD$. Moreover, the lower-resolution sentinel imagery is inherently blurry offering some regularization effect during training, leading to improved generalizability of our models. Following similar trends as in Table~\ref{table:1}, an absolute Recall@100 improvement of about $10$ points is observed, when using a pre-trained frozen audio encoder from L-CLAP. Similarly, the retrieval performance improves to $0.396$ when the audio encoder is allowed to be fine-tuned. We also observe gain in performance of fine-tuned GeoCLAP models trained with text containing address. The best performance for GeoCLAP trained with all three modalities, yields (Recall@100, Median Rank) of ($0.450$,$143$) and ($0.447$,$144$) for image-to-sound and sound-to-image retrieval, respectively. Compared to the baseline, this is a relative gain of $55.71\%$ and $57.95\%$ for Recall@100 on tasks: image-to-sound and sound-to-image retrieval, respectively.

\begin{table}
\scriptsize
 \begin{center}
    \begin{tabular}{ c c c c c|c c|c c  }
 \hline
 \multicolumn{5}{c|}{Method} &
 \multicolumn{2}{c|}{Image2Sound} &
 \multicolumn{2}{c}{Sound2Image} \\
 
 \hline
 Experiment & Image Encoder & Text-Audio Encoder & Text & Address & R@100 & Median-R & R@100 & Median-R\\
 \hline
 Baseline~\cite{heidler2023self} & ResNet18 & ResNet18 & \xmark &  \xmark & 0.289 & 620 & 0.283 & 635\\
 \hline
 ours & SATMAE & L-CLAP-frozen & \xmark &  \xmark &0.384 & 274 & 0.381 & 271\\
  ours & SATMAE & L-CLAP-frozen & \cmark &   \xmark & 0.340 & 369 & 0.338 & 367 \\
   ours & SATMAE & L-CLAP-frozen & \xmark &   \cmark &0.311 & 453 & 0.304 & 461\\
 ours & SATMAE & L-CLAP-frozen & \cmark &   \cmark &0.337 & 378 & 0.331 & 370\\
 \hline
 ours & SATMAE & L-CLAP & \xmark &   \xmark & 0.396 & 199 & 0.396 & 205\\
  ours & SATMAE & L-CLAP & \cmark &  \xmark & 0.441 & 152  & 0.441  & 155\\
  ours & SATMAE & L-CLAP & \xmark &  \cmark & 0.441 &153 &0.440 &156\\
 ours & SATMAE & L-CLAP & \cmark &  \cmark & \textbf{0.450} & \textbf{143} & \textbf{0.447} & \textbf{144}\\
\hline

\end{tabular}
\end{center}
  \caption{Cross-modal retrieval performance for models using 10m GSD overhead imagery.}
\label{table:2}
\end{table}

\subsection{Zero-Shot Soundscape Mapping}
\begin{figure}[!ht]
    \centering
    \includegraphics[width=130mm, height=45mm]{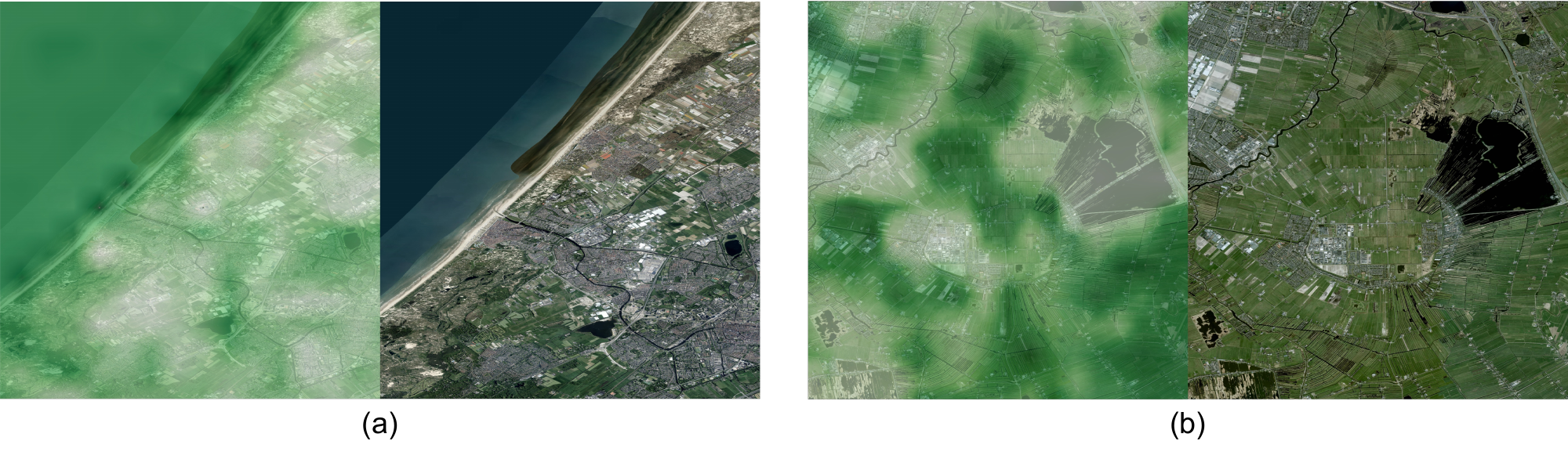}
    \caption{Soundscape maps along with reference overhead image for two regions. Soundscape created for queries:~(a) A textual prompt:~\textit{This is a sound of sea waves}; (b) randomly selected sound from the class \texttt{chirping\_birds} from ESC50 database~\cite{piczak2015dataset}~(green: more probable, white: less probable).}
    \label{fig:small_regions}
\end{figure}

\begin{figure}[!ht]
    \centering
    \includegraphics[width=130mm, height=60mm]{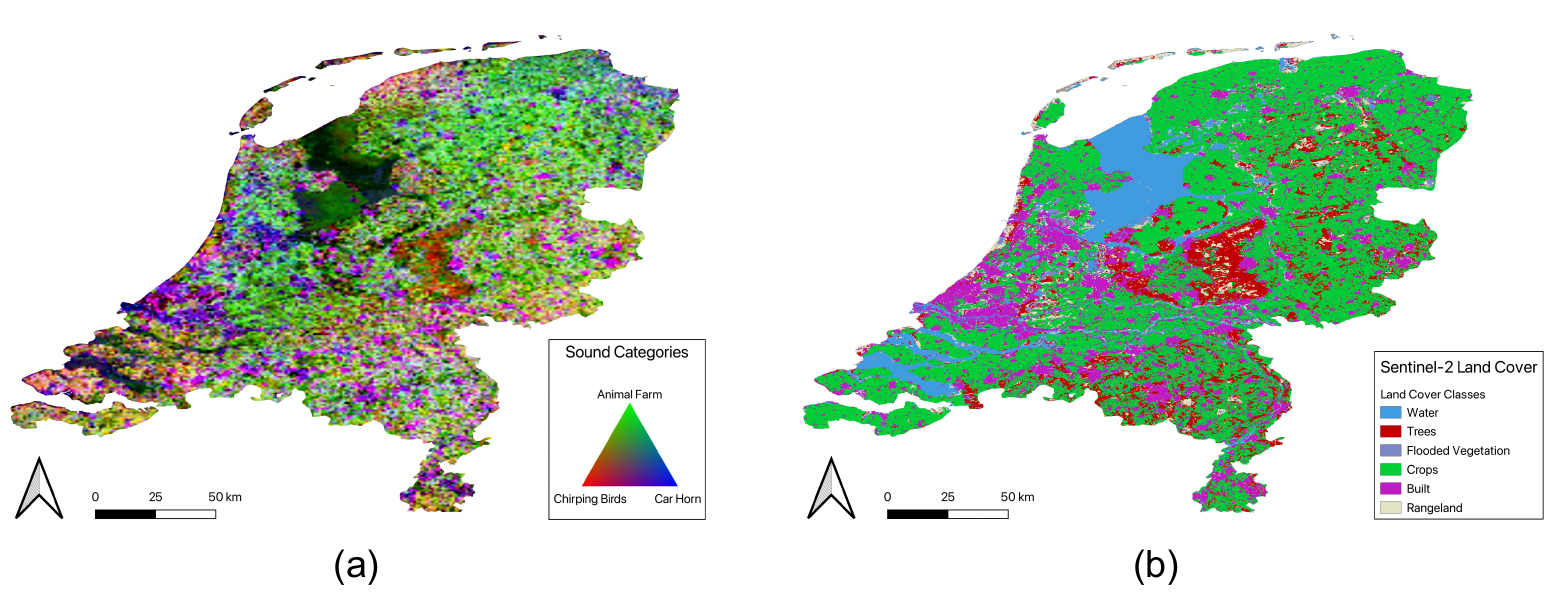}
    \caption{Comparison of~(a)~Soundscape map of the Netherlands with (b)~\textit{Sentinel-2 land cover classes}. The soundscape map was created by querying GeoCLAP with textual prompts for three sound categories: \textit{car horn}, \textit{chirping birds}, and \textit{animal farm}.}
    \label{fig:netherlands}
\end{figure}

Utilising the rich representation space of our best-performing GeoCLAP model, we demonstrate zero-shot soundscape mapping using both text and audio queries. Soundscape maps, in our work, are the similarity-score heatmaps for a given query. 
Specifically, we use the appropriate encoder from GeoCLAP to produce an embedding of the query and embeddings for a dense set of overhead images in the region of interest.
Then, the cosine similarity score between the query embedding and all overhead image embeddings is overlaid on the corresponding region to yield a soundscape map~(Figure~\ref{fig:small_regions}). In Figure~\ref{fig:netherlands}, we demonstrate a country-scale soundscape map for the Netherlands. For this, we compute soundscape for three prompts: \{\textit{This is a sound of car horn; This is a sound of chirping birds; This is a sound of animal farm}\} and overlay them together to create a composite pseudo-color map. We compare this soundscape with ESRI's \textit{Sentinel-2 land cover} classes. We observe a strikingly high correlation between the related land-cover classes with the category of sound likely to be heard at the location. More such soundscape maps can be found in the supplemental material of this paper.

\section{Conclusion}
\label{sec:conclusion}
We proposed GeoCLAP, a contrastive-learning framework capable of embedding the modalities of overhead imagery, audio, and text into a common space. Our approach significantly improves the state of the art for cross-modal retrieval between overhead imagery and audio. We utilise the learned, multi-modal representation space for soundscape mapping, demonstrating a simple and scalable way to create soundscape maps for any geographic area using only satellite imagery and audio or textual queries. With this approach, we can construct global, high-resolution soundmaps with minimal effort.

\bibliography{longstrings,ms}

\begin{thebibliography}{41}
\providecommand{\natexlab}[1]{#1}
\providecommand{\url}[1]{\texttt{#1}}
\expandafter\ifx\csname urlstyle\endcsname\relax
  \providecommand{\doi}[1]{doi: #1}\else
  \providecommand{\doi}{doi: \begingroup \urlstyle{rm}\Url}\fi

\bibitem[Aiello et~al.(2016)Aiello, Schifanella, Quercia, and Aletta]{aiello2016chatty}
Luca~Maria Aiello, Rossano Schifanella, Daniele Quercia, and Francesco Aletta.
\newblock Chatty maps: constructing sound maps of urban areas from social media data.
\newblock \emph{Royal Society open science}, 3\penalty0 (3):\penalty0 150690, 2016.

\bibitem[Aporee()]{aporee}
Radio Aporee.
\newblock https://aporee.org/maps.

\bibitem[Aramaki and Wakamiya(2023)]{aramaki2023image}
Eiji Aramaki and Shoko Wakamiya.
\newblock Image and sound of the city.
\newblock In \emph{The Social City: Space as Collaborative Media to Enhance the Value of the City}, pages 205--214. Springer, 2023.

\bibitem[Ayush et~al.(2021)Ayush, Uzkent, Meng, Tanmay, Burke, Lobell, and Ermon]{ayush2021geography}
Kumar Ayush, Burak Uzkent, Chenlin Meng, Kumar Tanmay, Marshall Burke, David Lobell, and Stefano Ermon.
\newblock Geography-aware self-supervised learning.
\newblock In \emph{Proceedings of the IEEE/CVF International Conference on Computer Vision}, 2021.

\bibitem[Chen et~al.(2022)Chen, Du, Zhu, Ma, Berg-Kirkpatrick, and Dubnov]{chen2022hts}
Ke~Chen, Xingjian Du, Bilei Zhu, Zejun Ma, Taylor Berg-Kirkpatrick, and Shlomo Dubnov.
\newblock Hts-at: A hierarchical token-semantic audio transformer for sound classification and detection.
\newblock In \emph{Proceedings of the IEEE International Conference on Acoustics, Speech and Signal Processing}, 2022.

\bibitem[Chen et~al.(2020)Chen, Lu, and Wang]{chen2020deep}
Yaxiong Chen, Xiaoqiang Lu, and Shuai Wang.
\newblock Deep cross-modal image--voice retrieval in remote sensing.
\newblock \emph{IEEE Transactions on Geoscience and Remote Sensing}, 58\penalty0 (10):\penalty0 7049--7061, 2020.

\bibitem[Cong et~al.(2022)Cong, Khanna, Meng, Liu, Rozi, He, Burke, Lobell, and Ermon]{cong2022satmae}
Yezhen Cong, Samar Khanna, Chenlin Meng, Patrick Liu, Erik Rozi, Yutong He, Marshall Burke, David Lobell, and Stefano Ermon.
\newblock Satmae: Pre-training transformers for temporal and multi-spectral satellite imagery.
\newblock \emph{Advances in Neural Information Processing Systems}, 35:\penalty0 197--211, 2022.

\bibitem[Cui et~al.(2022)Cui, Li, Xia, and Dai]{cui2022research}
Peng Cui, Tingting Li, Zhengwei Xia, and Chunyu Dai.
\newblock Research on the effects of soundscapes on human psychological health in an old community of a cold region.
\newblock \emph{International Journal of Environmental Research and Public Health}, 19\penalty0 (12):\penalty0 7212, 2022.

\bibitem[Deshmukh et~al.(2022)Deshmukh, Elizalde, and Wang]{deshmukh2022audio}
Soham Deshmukh, Benjamin Elizalde, and Huaming Wang.
\newblock Audio retrieval with wavtext5k and clap training.
\newblock \emph{arXiv preprint arXiv:2209.14275}, 2022.

\bibitem[Devlin et~al.(2018)Devlin, Chang, Lee, and Toutanova]{devlin2018bert}
Jacob Devlin, Ming-Wei Chang, Kenton Lee, and Kristina Toutanova.
\newblock Bert: Pre-training of deep bidirectional transformers for language understanding.
\newblock In \emph{Proceedings of the Conference of the North American Chapter of the Association for Computational Linguistics}, 2018.

\bibitem[Dosovitskiy et~al.(2021)Dosovitskiy, Beyer, Kolesnikov, Weissenborn, Zhai, Unterthiner, Dehghani, Minderer, Heigold, Gelly, Uszkoreit, and Houlsby]{dosovitskiy2020vit}
Alexey Dosovitskiy, Lucas Beyer, Alexander Kolesnikov, Dirk Weissenborn, Xiaohua Zhai, Thomas Unterthiner, Mostafa Dehghani, Matthias Minderer, Georg Heigold, Sylvain Gelly, Jakob Uszkoreit, and Neil Houlsby.
\newblock An image is worth 16x16 words: Transformers for image recognition at scale.
\newblock \emph{Proceedings of the International Conference on Learning Representations}, 2021.

\bibitem[Elizalde et~al.(2023)Elizalde, Deshmukh, Al~Ismail, and Wang]{elizalde2023clap}
Benjamin Elizalde, Soham Deshmukh, Mahmoud Al~Ismail, and Huaming Wang.
\newblock Clap learning audio concepts from natural language supervision.
\newblock In \emph{Proceedings of the IEEE International Conference on Acoustics, Speech and Signal Processing}, 2023.

\bibitem[Engel et~al.(2021)Engel, Fiebig, Pfaffenbach, and Fels]{engel2021review}
Margret~Sibylle Engel, Andr{\'e} Fiebig, Carmella Pfaffenbach, and Janina Fels.
\newblock A review of the use of psychoacoustic indicators on soundscape studies.
\newblock \emph{Current Pollution Reports}, pages 1--20, 2021.

\bibitem[for Standardization(2014)]{international2014iso}
International~Organization for Standardization.
\newblock Iso 12913-1: 2014 acoustics—soundscape—part 1: definition and conceptual framework.
\newblock \emph{ISO, Geneva}, 2014.

\bibitem[Garz{\'o}n et~al.(2023)Garz{\'o}n, Bravo-Moncayo, Arellana, and de~Dios~Ort{\'u}zar]{garzon2023relationships}
Luis Garz{\'o}n, Luis Bravo-Moncayo, Juli{\'a}n Arellana, and Juan de~Dios~Ort{\'u}zar.
\newblock On the relationships between auditory and visual factors in a residential environment context: A sem approach.
\newblock \emph{Frontiers in Psychology}, 14, 2023.

\bibitem[Gonz{\'a}lez et~al.(2023)Gonz{\'a}lez, Morillas, and Rey-Gozalo]{gonzalez2023effects}
David~Montes Gonz{\'a}lez, Juan Miguel~Barrig{\'o}n Morillas, and Guillermo Rey-Gozalo.
\newblock Effects of noise on pedestrians in urban environments where road traffic is the main source of sound.
\newblock \emph{Science of the total environment}, 857:\penalty0 159406, 2023.

\bibitem[Guzhov et~al.(2022)Guzhov, Raue, Hees, and Dengel]{guzhov2022audioclip}
Andrey Guzhov, Federico Raue, J{\"o}rn Hees, and Andreas Dengel.
\newblock Audioclip: Extending clip to image, text and audio.
\newblock In \emph{Proceedings of the IEEE International Conference on Acoustics, Speech and Signal Processing}, 2022.

\bibitem[He et~al.(2022)He, Chen, Xie, Li, Doll{\'a}r, and Girshick]{he2022masked}
Kaiming He, Xinlei Chen, Saining Xie, Yanghao Li, Piotr Doll{\'a}r, and Ross Girshick.
\newblock Masked autoencoders are scalable vision learners.
\newblock In \emph{Proceedings of the IEEE/CVF Conference on Computer Vision and Pattern Recognition}, 2022.

\bibitem[Heidler et~al.(2023)Heidler, Mou, Hu, Jin, Li, Gan, Wen, and Zhu]{heidler2023self}
Konrad Heidler, Lichao Mou, Di~Hu, Pu~Jin, Guangyao Li, Chuang Gan, Ji-Rong Wen, and Xiao~Xiang Zhu.
\newblock Self-supervised audiovisual representation learning for remote sensing data.
\newblock \emph{International Journal of Applied Earth Observation and Geoinformation}, 116:\penalty0 103130, 2023.

\bibitem[Hu et~al.(2020)Hu, Li, Mou, Jin, Chen, Jing, Zhu, and Dou]{hu2020cross}
Di~Hu, Xuhong Li, Lichao Mou, Pu~Jin, Dong Chen, Liping Jing, Xiaoxiang Zhu, and Dejing Dou.
\newblock Cross-task transfer for geotagged audiovisual aerial scene recognition.
\newblock In \emph{Proceedings of the European Conference on Computer Vision}. Springer, 2020.

\bibitem[Lercher and Dzhambov(2023)]{lercher2023soundscape}
Peter Lercher and Angel~M Dzhambov.
\newblock Soundscape and health.
\newblock In \emph{Soundscapes: Humans and Their Acoustic Environment}, pages 243--276. Springer, 2023.

\bibitem[Lionello et~al.(2020)Lionello, Aletta, and Kang]{lionello2020systematic}
Matteo Lionello, Francesco Aletta, and Jian Kang.
\newblock A systematic review of prediction models for the experience of urban soundscapes.
\newblock \emph{Applied Acoustics}, 170:\penalty0 107479, 2020.

\bibitem[Liu et~al.(2019)Liu, Ott, Goyal, Du, Joshi, Chen, Levy, Lewis, Zettlemoyer, and Stoyanov]{liu2019roberta}
Yinhan Liu, Myle Ott, Naman Goyal, Jingfei Du, Mandar Joshi, Danqi Chen, Omer Levy, Mike Lewis, Luke Zettlemoyer, and Veselin Stoyanov.
\newblock Roberta: A robustly optimized bert pretraining approach.
\newblock \emph{arXiv preprint arXiv:1907.11692}, 2019.

\bibitem[Liu et~al.(2021)Liu, Lin, Cao, Hu, Wei, Zhang, Lin, and Guo]{liu2021swin}
Ze~Liu, Yutong Lin, Yue Cao, Han Hu, Yixuan Wei, Zheng Zhang, Stephen Lin, and Baining Guo.
\newblock Swin transformer: Hierarchical vision transformer using shifted windows.
\newblock In \emph{Proceedings of the IEEE/CVF International Conference on Computer Vision}, 2021.

\bibitem[Mao et~al.(2018)Mao, Yuan, and Xiaoqiang]{mao2018deep}
Guo Mao, Yuan Yuan, and Lu~Xiaoqiang.
\newblock Deep cross-modal retrieval for remote sensing image and audio.
\newblock In \emph{10th IAPR workshop on pattern recognition in remote sensing}, 2018.

\bibitem[Margaritis and Kang(2017)]{margaritis2017soundscape}
Efstathios Margaritis and Jian Kang.
\newblock Soundscape mapping in environmental noise management and urban planning: case studies in two uk cities.
\newblock \emph{Noise mapping}, 4\penalty0 (1):\penalty0 87--103, 2017.

\bibitem[Ning et~al.(2021)Ning, Zhao, and Yuan]{ning2021semantics}
Hailong Ning, Bin Zhao, and Yuan Yuan.
\newblock Semantics-consistent representation learning for remote sensing image--voice retrieval.
\newblock \emph{IEEE Transactions on Geoscience and Remote Sensing}, 60:\penalty0 1--14, 2021.

\bibitem[Ooi et~al.(2023)Ooi, Ong, Watcharasupat, Lam, Hong, and Gan]{ooi2023araus}
Kenneth Ooi, Zhen-Ting Ong, Karn~N Watcharasupat, Bhan Lam, Joo~Young Hong, and Woon-Seng Gan.
\newblock Araus: A large-scale dataset and baseline models of affective responses to augmented urban soundscapes.
\newblock \emph{IEEE Transactions on Affective Computing}, 2023.

\bibitem[Owens et~al.(2016)Owens, Wu, McDermott, Freeman, and Torralba]{owens2016ambient}
Andrew Owens, Jiajun Wu, Josh~H McDermott, William~T Freeman, and Antonio Torralba.
\newblock Ambient sound provides supervision for visual learning.
\newblock In \emph{Proceedings of the European Conference on Computer Vision}, 2016.

\bibitem[Picaut et~al.(2019)Picaut, Fortin, Bocher, Petit, Aumond, and Guillaume]{picaut2019open}
Judica{\"e}l Picaut, Nicolas Fortin, Erwan Bocher, Gwendall Petit, Pierre Aumond, and Gwena{\"e}l Guillaume.
\newblock An open-science crowdsourcing approach for producing community noise maps using smartphones.
\newblock \emph{Building and Environment}, 148:\penalty0 20--33, 2019.

\bibitem[Piczak(2015)]{piczak2015dataset}
Karol~J. Piczak.
\newblock {ESC}: {Dataset} for {Environmental Sound Classification}.
\newblock In \emph{Proceedings of the Association for Computing Machinery Conference on Multimedia}, 2015.

\bibitem[Radford et~al.(2021)Radford, Kim, Hallacy, Ramesh, Goh, Agarwal, Sastry, Askell, Mishkin, Clark, et~al.]{radford2021learning}
Alec Radford, Jong~Wook Kim, Chris Hallacy, Aditya Ramesh, Gabriel Goh, Sandhini Agarwal, Girish Sastry, Amanda Askell, Pamela Mishkin, Jack Clark, et~al.
\newblock Learning transferable visual models from natural language supervision.
\newblock In \emph{International Conference on Machine Learning}. PMLR, 2021.

\bibitem[Radicchi et~al.(2021)Radicchi, Cevikayak~Yelmi, Chung, Jordan, Stewart, Tsaligopoulos, McCunn, and Grant]{radicchi2021sound}
Antonella Radicchi, P{\i}nar Cevikayak~Yelmi, Andy Chung, Pamela Jordan, Sharon Stewart, Aggelos Tsaligopoulos, Lindsay McCunn, and Marcus Grant.
\newblock Sound and the healthy city.
\newblock \emph{Cities \& Health}, 5\penalty0 (1-2):\penalty0 1--13, 2021.

\bibitem[Reed et~al.(2022)Reed, Gupta, Li, Brockman, Funk, Clipp, Candido, Uyttendaele, and Darrell]{reed2022scale}
Colorado~J Reed, Ritwik Gupta, Shufan Li, Sarah Brockman, Christopher Funk, Brian Clipp, Salvatore Candido, Matt Uyttendaele, and Trevor Darrell.
\newblock Scale-mae: A scale-aware masked autoencoder for multiscale geospatial representation learning.
\newblock \emph{arXiv preprint arXiv:2212.14532}, 2022.

\bibitem[Salem et~al.(2018)Salem, Zhai, Workman, and Jacobs]{salem2018multimodal}
Tawfiq Salem, Menghua Zhai, Scott Workman, and Nathan Jacobs.
\newblock A multimodal approach to mapping soundscapes.
\newblock In \emph{Proceedings of the IEEE Conference on Computer Vision and Pattern Recognition Workshops}, 2018.

\bibitem[Wang et~al.(2022)Wang, Albrecht, Braham, Mou, and Zhu]{wang2022self}
Yi~Wang, Conrad Albrecht, Nassim Ait~Ali Braham, Lichao Mou, and Xiaoxiang Zhu.
\newblock Self-supervised learning in remote sensing: A review.
\newblock \emph{IEEE Geoscience and Remote Sensing Magazine}, 2022.

\bibitem[Wu et~al.(2022)Wu, Seetharaman, Kumar, and Bello]{wu2022wav2clip}
Ho-Hsiang Wu, Prem Seetharaman, Kundan Kumar, and Juan~Pablo Bello.
\newblock Wav2clip: Learning robust audio representations from clip.
\newblock In \emph{Proceedings of the IEEE International Conference on Acoustics, Speech and Signal Processing}, 2022.

\bibitem[Wu* et~al.(2023)Wu*, Chen*, Zhang*, Hui*, Berg-Kirkpatrick, and Dubnov]{laionclap2023}
Yusong Wu*, Ke~Chen*, Tianyu Zhang*, Yuchen Hui*, Taylor Berg-Kirkpatrick, and Shlomo Dubnov.
\newblock Large-scale contrastive language-audio pretraining with feature fusion and keyword-to-caption augmentation.
\newblock In \emph{Proceedings of the IEEE International Conference on Acoustics, Speech and Signal Processing}, 2023.

\bibitem[Yang et~al.(2022)Yang, Wang, Sun, Zhang, Liao, Gu, Hou, and Jiao]{yang2022multimodal}
Rui Yang, Shuang Wang, Yingzhi Sun, Huan Zhang, Yu~Liao, Yu~Gu, Biao Hou, and Licheng Jiao.
\newblock Multimodal fusion remote sensing image--audio retrieval.
\newblock \emph{IEEE Journal of Selected Topics in Applied Earth Observations and Remote Sensing}, 15:\penalty0 6220--6235, 2022.

\bibitem[Yue et~al.(2023)Yue, Meng, Yang, Wu, Liu, and Yan]{yue2023visualized}
Ran Yue, Qi~Meng, Da~Yang, Yue Wu, Fangfang Liu, and Wei Yan.
\newblock A visualized soundscape prediction model for design processes in urban parks.
\newblock In \emph{Building Simulation}, volume~16, pages 337--356. Springer, 2023.

\bibitem[Zhao et~al.(2023)Zhao, Liang, Tu, Huang, and Biljecki]{zhao2023sensing}
Tianhong Zhao, Xiucheng Liang, Wei Tu, Zhengdong Huang, and Filip Biljecki.
\newblock Sensing urban soundscapes from street view imagery.
\newblock \emph{Computers, Environment and Urban Systems}, 99:\penalty0 101915, 2023.

\end{thebibliography}
\end{document}


\maketitle

In this supplemental material, we present a demonstration of the zero-shot soundscape mapping capability offered by our proposed framework, GeoCLAP. Specifically, we showcase the soundscape maps created by querying our best performing model with diverse sound-related textual prompts. Furthermore, in a \href{https://drive.google.com/file/d/1ljBnTuuQXB-GZkA3NcingScfNy72QFIY/view?usp=sharing}{video demonstration} accompanying this material, we highlight the satellite image to audio retrieval capability of GeoCLAP.

\section{Zero-Shot Soundscape Mapping}
\label{sec:intro}
Following the same methodology from Section $5.3$ in the main paper, we constructed a soundscape map of England. We selected three prompts: \textit{This is a sound of car horn; This is a sound of chirping birds; This is a sound of animal farm}. We downloaded Sentinel-2 cloudless images for England, each with dimension $256 \times 256$. Then, using cosine similarity scores between image and text embeddings, we created a dense soundscape map for the region. All visualizations were created using Q-GIS.

\begin{figure}[!htb]
    \centering
    \includegraphics[width=120mm, height=65mm]{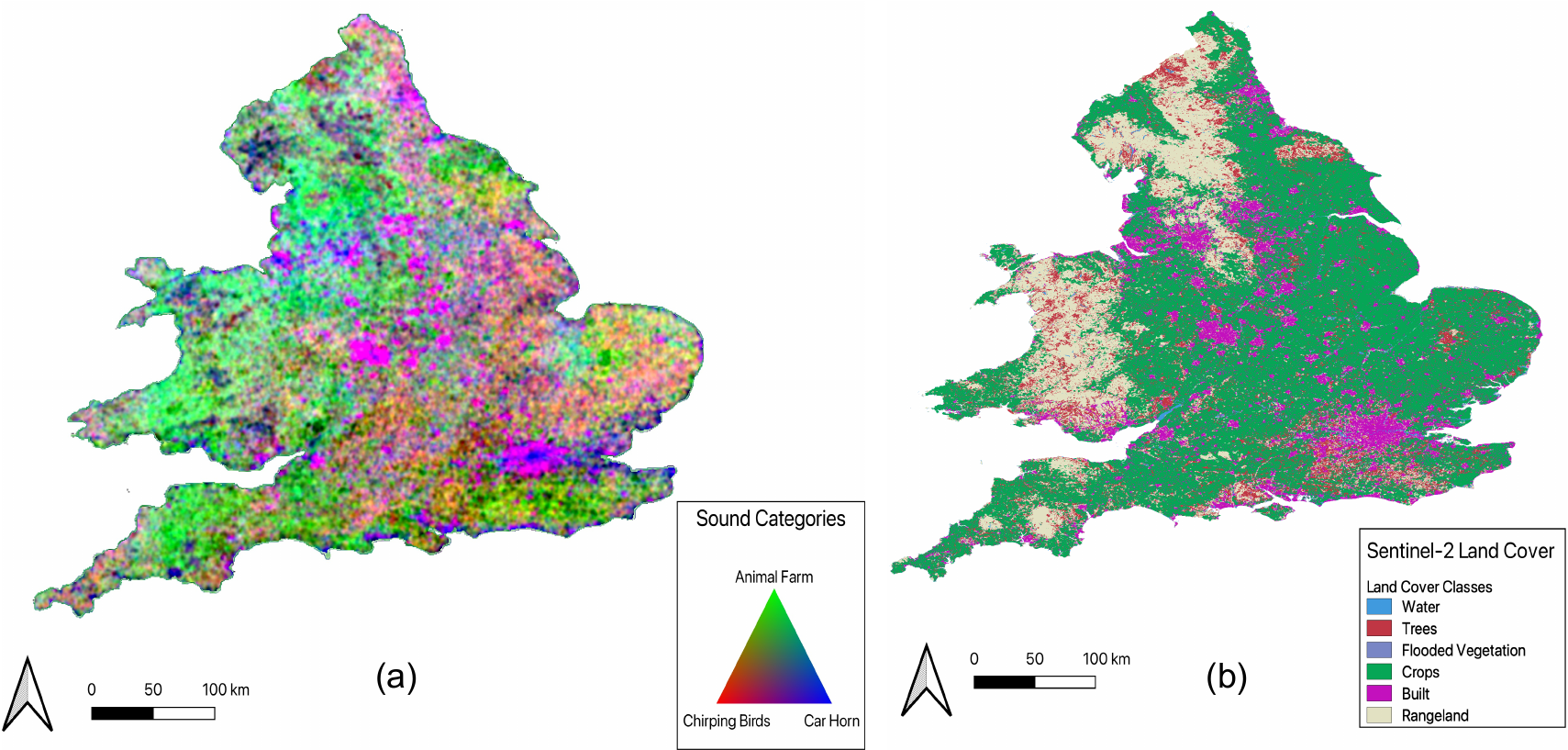}
    \caption{Comparison of~(a)~Soundscape map of  England with (b)~\textit{ESRI's Sentinel-2 land cover classes}. The soundscape map was created by querying GeoCLAP with textual prompts for three sound categories: \textit{car horn}, \textit{chirping birds}, and \textit{animal farm}. Best viewed in colour.}
    \label{fig:england}
\end{figure}

As observed in Figure \ref{fig:england}, there is a strong correlation between sound categories and relevant land-cover classes. As expected, the soundscape map reveals that urban areas in England, such as the region around London, are highly associated with the sound category \textit{car horn} indicated by the colour blue in Figure \ref{fig:england}~(a). On the other hand, less populated areas with crops exhibit a notable association with the sound category \textit{animal farm}. An intriguing observation is that around built-up areas in England, a combination of both \textit{car horn} and \textit{chirping birds} sound is observed, as indicated by purple-coloured regions in soundscape. This suggests that despite human activities in these regions, birds still inhabit them.

\begin{figure}[!htb]
    \centering
    \includegraphics[width=125mm, height=55mm]{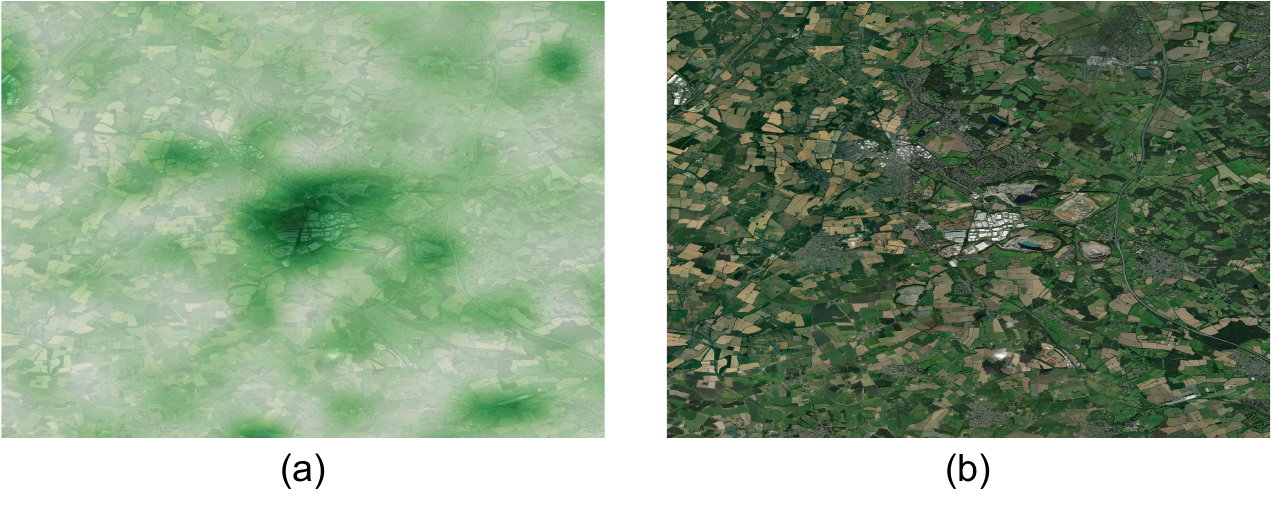}
    \caption{Soundscape map of a small region (a) along with the reference overhead image (b). Soundscape created for the textual prompt:~\textit{This is a sound of manufacturing factory}.~(green: more probable, white: less probable).}
    \label{fig:factory}
\end{figure}

\begin{figure}[!htb]
    \centering
    \includegraphics[width=125mm, height=55mm]{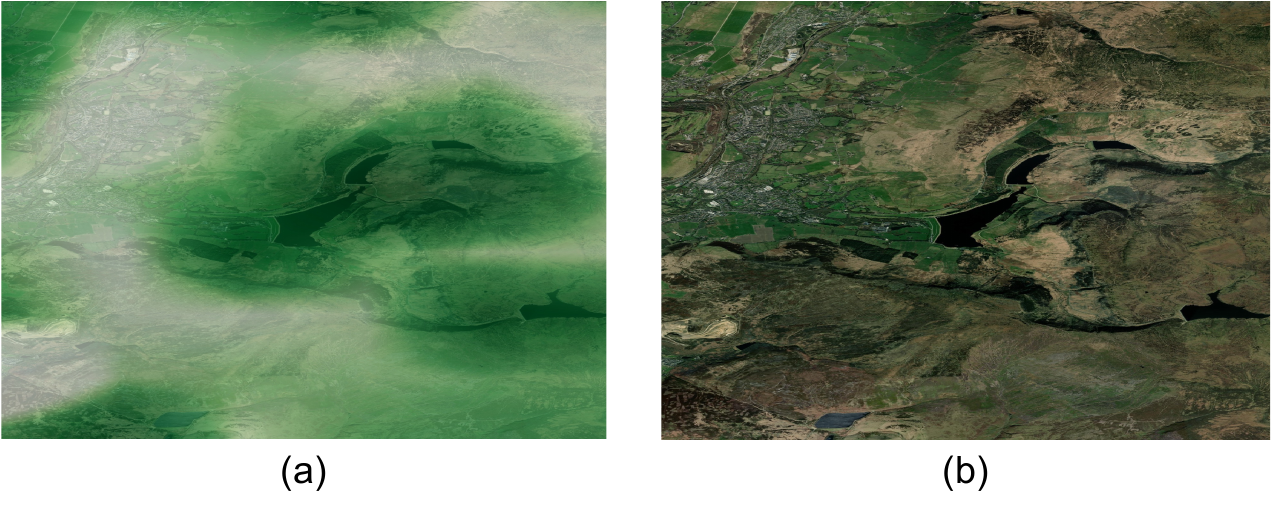}
    \caption{Soundscape map of a small region (a) along with the reference overhead image (b). Soundscape created for the textual prompt:~\textit{This is a sound of flowing river}.~(green: more probable, white: less probable).}
    \label{fig:river}
\end{figure}

\begin{figure}[!htb]
    \centering
    \includegraphics[width=75mm, height=75mm]{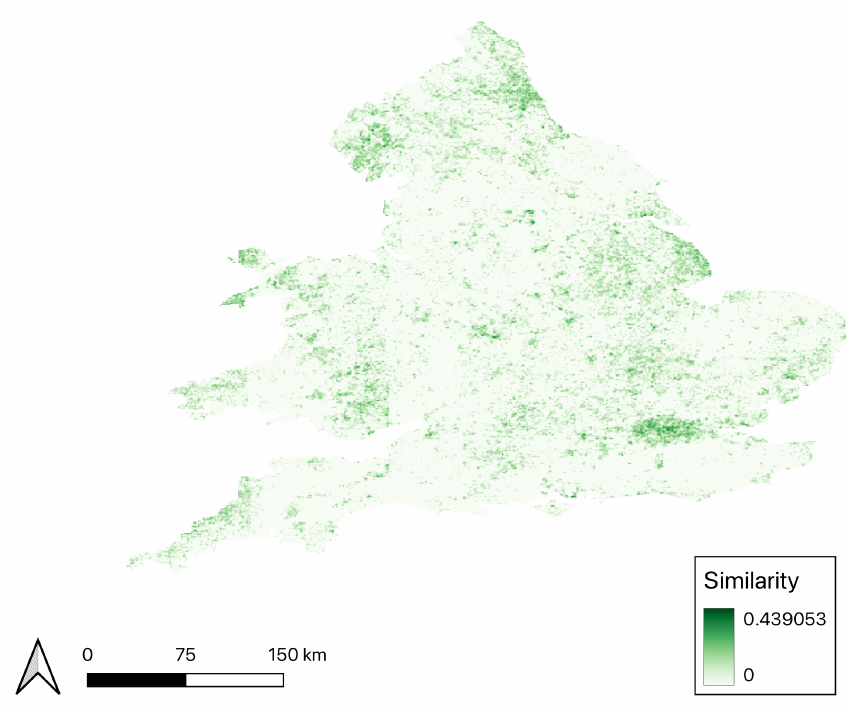}
    \caption{Soundscape map of England created by querying GeoCLAP with a textual prompt: \textit{This is a sound of church bells}.}
    \label{fig:church_bells}
\end{figure}

Soundscapes can be viewed as composite pseudo-colour maps representing a desired set of sound categories, as shown in Figure \ref{fig:england}. However, if one is specifically interested in a single sound category, the GeoCLAP model can be queried with a textual prompt corresponding to that particular sound category, as demonstrated in Figure \ref{fig:church_bells}. Furthermore, visualizing soundscapes for smaller geographic regions, as showcased in Figure \ref{fig:factory} and \ref{fig:river}, can provide a better understanding of sound-related concepts learned by the model.

The results shown in Figure \ref{fig:factory} indicate high similarity between the prompt:~\textit{This is a sound of a manufacturing factory} and a sub-region that likely contains structures resembling manufacturing factories. Similarly, in Figure \ref{fig:river}, areas associated with water bodies exhibit a high similarity with the prompt:~\textit{This is a sound of a flowing river.} These findings demonstrate that the embedding space of GeoCLAP possesses an understanding of high-level sound-related concepts within geographic regions.



